\documentclass[sn-basic]{sn-jnl}

\usepackage{graphicx}%
\usepackage{multirow}%
\usepackage{amsmath,amssymb,amsfonts}%
\usepackage{amsthm}%
\usepackage{mathrsfs}%
\usepackage[title]{appendix}%
\usepackage{xcolor}%
\usepackage{textcomp}%
\usepackage{manyfoot}%
\usepackage{booktabs}%
\usepackage{booktabs}%
\usepackage{url}
\usepackage{algorithm}%
\usepackage{algorithmicx}%
\usepackage{algpseudocode}%
\usepackage{listings}%
\usepackage{tikz}
\usepackage{pgfplots}
\pgfplotsset{compat=1.17}
\usepackage{pgfplotstable}
\usetikzlibrary{patterns,shapes,arrows.meta,calc,positioning}
\usepackage{sansmath}
\usepackage{float}
\usepackage[utf8]{inputenc}
\renewcommand{\thetable}{\Roman{table}}

\theoremstyle{thmstyleone}%
\theoremstyle{thmstyletwo}%

\theoremstyle{thmstylethree}%
\newcommand{\dataset}{\mathcal{D}}   
\newcommand{\reals}{\mathbb{R}}      

\usepackage[framemethod=TikZ]{mdframed}

\newmdenv[
  linewidth=0.8pt,
  roundcorner=5pt,
  backgroundcolor=gray!5,
  linecolor=black,
  skipabove=10pt,
  skipbelow=10pt,
  innertopmargin=8pt,
  innerbottommargin=8pt,
  leftmargin=0pt,
  rightmargin=0pt
]{springerbox}

\newcommand{\definitionbox}[2]{%
  \begin{springerbox}
    \textbf{#1.} #2
  \end{springerbox}
}

\begin{document}

\title[Future of AI Models: A Computational perspective on Model collapse]{Future of AI Models: A Computational perspective on Model collapse}


\author*[1]{\fnm{Trivikram} \sur{Satharasi}}\email{t.satharasi@ufl.edu}

\author[2]{\fnm{S. Sitharama} \sur{Iyengar}}\email{iyengar@cis.fiu.edu}

\affil*[1]{\orgname{University of Florida}, \orgaddress{\city{Gainesville}, \postcode{32611}, \state{Florida}, \country{United States}}}

\affil[2]{\orgname{Florida International University}, \orgaddress{\city{Miami}, \postcode{33199}, \state{Florida}, \country{United States}}}


\abstract{In recent years, Artificial Intelligence, especially with the advent of Large Language Models (LLMs), has transformed domains\citep{naveed2025comprehensive} such as software engineering, journalism, and creative writing, and is now reshaping academia and media through diffusion models like Stable Diffusion, which generate high-quality images and videos from text. Recent evidence illustrates this growing trend, A 2025 Ahrefs study found that 74.2\% of newly published webpages contain AI-generated material \citep{ahrefs2025}, while large-scale text analyses estimate that 30–40\% of the active web corpus is now synthetic \citep{arxiv2025contamination} and also a survey \citep{elon2025survey} have show that 52\% of U.S. adults use LLMs such as ChatGPT for writing, coding, or research tasks, and institutional audits reveal that 18\% of financial complaint texts and 24\% of corporate press releases are AI-assisted \citep{arxiv2025pressdata}.

The neural architectures powering these systems, for example, Transformers, RNNs, LSTMs, GANs, and diffusion networks, depend critically on large, diverse, human-authored datasets\citep{DiscoveryLab.ShiIyengar2020,vaswani2023attentionneed}. As AI-assisted generation becomes the norm for many tasks, the share of synthetic (AI-generated) content relative to human-authored data continues to rise, creating a recursive feedback loop that risks eroding linguistic and semantic diversity—a process known as \textbf{\textit{Model Collapse}}\citep{shumailov2024collapse,dohmatob2024strongmodelcollapse}.

This study quantifies and forecasts the potential onset of model collapse by examining the evolution of linguistic similarity in large-scale text datasets. Using a filtered subset of the Common Crawl corpus (English-language Wikipedia articles), we compute year-wise semantic similarity from 2013 to 2025 through Transformer-based embeddings and cosine metrics. Results indicate a steady rise in similarity even before public LLM adoption, likely due to early use of RNNs and LSTMs for translation and text normalization. However, their limited scale kept the effect modest alongside. Results also indicate fluctuations in similarity are attributed to irreducible linguistic diversity, variations in corpus size across years and finite sampling error. These findings provide a data-driven estimate for when recursive AI contamination could significantly threaten data richness and model generalization.}




\maketitle
\section{Introduction}\label{sec:intro}
The recent years have seen Artificial Intelligence undergo a paradigm shift, largely driven by sophisticated generative and language models. Large Language Models (LLMs) \citep{naveed2025comprehensive} like OpenAI’s GPT series \citep{gpt2,gpt3,openai2024gpt4technicalreport} have fundamentally transformed business models of various industries. These models have demonstrated emergent capabilities that scale with size. They are no longer mere tools but collaborative partners, automating complex cognitive tasks like software development, legal contract analysis, and scientific research. Simultaneously, they serve as powerful catalysts for human creativity, assisting authors in drafting narratives, helping marketers brainstorm campaigns, and enabling developers to prototype ideas at unprecedented speed. These capabilities have been discussed in greater detail in \citep{naveed2025comprehensive}.

Parallel to the advances in language generation, visual media capabilities have experienced a similar revolution (For example OpenAI's DALLE \citep{dalle:ramesh2021zeroshottexttoimagegeneration,dalle:ramesh2022hierarchicaltextconditionalimagegeneration}), driven by text-to-image based models and further revolutionized by diffusion-based systems such as Stable Diffusion \citep{croitoru2023,rombach2022highresolutionimagesynthesislatent,ho2020denoisingdiffusionprobabilisticmodels}  . These models have reshaped professional workflows \citep{WANG2024127837,10.1145/3678884.3681890,tang2024exploringimpactaigeneratedimage} in graphic design, architecture, product prototyping, and entertainment, where concept artists and directors can now visualize scenes with ease and edit them with flexibility. The scale of this transformation is extraordinary: as of 2024, more than \textbf{15 billion AI-generated images} have been created using diffusion-based systems, with estimates suggesting over \textbf{30 million new images generated daily} \citep{journal_everypixel_2024,amitkumar2024_ai_images}. The training datasets enabling these systems, such as LAION-5B\citep{schuhmann2022laion5bopenlargescaledataset}, contain \textbf{billions of image–text pairs} scraped from the public web \citep{rombach2022highresolutionimagesynthesislatent}. A systematic review by Oksanen et al. (2023) \citep{oksanen2023finearts} analyzed 723 studies on AI in fine arts and retained 44 empirical works; over half focused on visual arts, while one quarter examined AI in music generation. Notably, several studies found that audiences often struggle to distinguish AI-generated from human-created artworks, though human art is still rated higher in perceived aesthetic depth. More recently, a meta-analysis by Holzner et al. (2025) \citep{holzner2025generativecreativity} aggregated data from \textbf{28 studies involving 8,214 participants}, showing that humans augmented with generative AI outperform unaided humans in creativity tasks (Hedges’ $g = 0.27$), while GenAI use substantially reduces idea diversity ($g = -0.86$), empirically validating concerns over homogenization.

The performance of these transformative techniques lies in their training techniques, which are critically dependent on vast, high-quality, and richly diverse datasets, most of which have been scraped from the public internet. These datasets are regarded as a comprehensive digital snapshot of collective human knowledge, language, logic, and creativity. This snapshot includes not only facts but also dialogue, the logical structure of code, the tone of a sentence/language, and literature. The diversity and factual accuracy of this data are the bedrock of AI models’ capabilities. It is this foundation that allows the models to achieve their generalization. However, these foundational datasets also carry misinformation and inherent biases. The future development of AI, therefore, hinges not just on bigger models but also on curation and filtering of the data that gives them life and accuracy of information. The curation and filtering of data during the modeling of large-scale AI systems are fundamental to achieving sustained accuracy and reliability over time. Raw datasets collected from heterogeneous sources often exhibit high levels of noise, redundancy, imbalance, and latent bias, which can severely degrade model performance and accelerate accuracy decay in dynamic environments. Systematic data preprocessing, through deduplication, normalization, feature selection, and outlier removal, enhances the signal-to-noise ratio and provides more stable statistical properties for downstream model training. Furthermore, bias-aware filtering and fairness-driven data augmentation mitigate representational disparities, thereby reducing systematic errors and improving model equity across diverse subpopulations. These measures are particularly critical in mitigating data drift and concept drift, where underlying data distributions shift over time due to evolving user behaviors, environmental factors, or adversarial manipulation.
Incorporating active learning and continual learning strategies further strengthens model resilience by dynamically updating training corpora with high-value samples while discarding low-utility or adversarial data points. Recent work in data-centric AI emphasizes\citep{DiscoveryLab.Iyengar2025.Convergence,DiscoveryLab.Iyengar2025.Future,DiscoveryLab.SingaramIyengarMadni2024,DiscoveryLab.SoniGurappaUpadhyay2024,NgEtAl2022.DataCentricAICompetition,DiscoveryLab.ThejasHariprasadIyengarSunithaPrajwalChennupati2022,DiscoveryLab.ShiIyengar2020,northcutt2021confident}, high-quality data governance frameworks not only optimize training efficiency but also extend the operational lifespan of AI models by preserving generalization capability in non-stationary environments. Ultimately, well-curated and adaptively filtered datasets serve as the cornerstone for scalable, trustworthy, and temporally robust AI systems. 

The effect of the proliferation of AI is seen in a  2025 study by Ahrefs reporting 74.2\% of newly published web pages contained AI-generated material \citep{ahrefs2025}. Large-scale web-corpus analyses estimate that 30–40\% of all active web text now originates from AI-generated or AI-edited sources \citep{arxiv2025contamination}. Furthermore, a 2025 survey by Elon University found that 52\% of U.S. adults regularly use LLMs like ChatGPT for writing, coding, or research tasks \citep{elon2025survey}. Within professional domains, 18\% of financial consumer complaint records and 24\% of corporate press releases are estimated to contain LLM-assisted text \citep{arxiv2025pressdata}.

This proliferation of AI-generated content has introduced a looming, systemic risk to the future of artificial intelligence itself: \textbf{AI Model Collapse}. This phenomenon, sometimes referred to in academic literature as \textbf{Model Autophagy Disorder}, describes a degenerative process where successive generations of models exhibit progressively diminishing quality when trained on the synthetic data produced by their predecessors. The very scalability that makes generative AI so powerful becomes its potential undoing, as the digital ecosystem becomes saturated with its own artificial creations, threatening to poison the well for future training by creating a self-consuming feedback loop.

In this work, a timeline for the potential collapse of AI models is analyzed and generated. This timeline is estimated by examining similarities between past and current data, taking into account the current rates of synthetic data consumption and availability. This work is divided into sections: Section \ref{sec:Problem_Statement} discusses the Problem Statement and its relevance; Section \ref{sec:Theoretical_Intuition} provides Theoretical Intuition; Section \ref{sec:Data} shows the Properties of the Analyzed data; Section \ref{sec:Discussion} discusses the Results and Intuitive Inferences; and Section \ref{sec:Conclusion} summarizes the results and Future Work in this area.

\section{Problem Statement}\label{sec:Problem_Statement}
This work aims to analyze the effect of the proliferation of Artificial Intelligence on the quality of datasets over time. Mathematically, given a corpus of data from a single source, this study quantifies the trends in similarity and diversity along the temporal axis. In particular, we investigate the temporal distribution, context, and writing style through similarity and overall diversity of the corpus as it evolves.

Let the corpus, denoted by \(\dataset\), consist of $N$ datasets, indexed by $i=1, 2,... N$. For document $i$, let 
\begin{itemize}
  \item \(y_i\ \in \reals\) denotes the year or order of creation,
  \item \(x_{i,j} \in \reals^m\) denotes the content representation of element $j$ in the dataset $\dataset[i|y=y_i]$ (e.g.\ embedding vector capturing style and context) 
\end{itemize}
We define the following parameters:
\begin{itemize}
  \item \(q_{i,j,k}\in \reals \) defines the  similarity between element $j$ and $k$ in the dataset $\dataset[i|y=y_i]$,
  \item \(q_{i}\in \reals \) defines the  average similarity between all elements in the dataset $\dataset[i|y=y_i]$
  $$ q_{i}= \frac{1}{\mathcal{N}_i}\sum_{j,k \in \dataset(y_i)} q_{i,j,k}$$ 
  where,\\
\subitem $\mathcal{N}$ is the number of distinct combinations $q_{i,j,k}$ for distinct $x_{i,j}$ and $x_{i,k}$ in the dataset $\dataset[i]$
  \item $\mu_{i}  \in \reals^m $ defines the mean of all the representations in a specific dataset,i.e., $x_{i,j} \in \dataset[i]$.
  where, \\
 \subitem $$\mu_{i}=\frac{1}{n_i}\sum_{j \in \dataset[i|y=y_i]} x_{i,j}$$ where $n_i$ is the no of elements in the dataset $\dataset[i]$
 \item $\Sigma_{i} \in \reals^{m\times m}$ defines the variance of all the representations in a specific dataset,i.e., $x_{i,j} \in \dataset[i]$.
  where, \\
 \subitem $$\Sigma_{i}=\frac{1}{n_i-1}\sum_{j \in \dataset[i|y=y_i]} \left[x_{i,j}-\mu_i\right]\left[x_{i,j}-\mu_i\right]^T$$ where $n_i$ is the no of elements in the dataset $\dataset[i]$
\end{itemize}

This work will mathematically visualize the evolution of similarity and diversity over time in the dataset, enabling us to detect and interpret any potential onset of collapse or excessive redundancy.

\section{Theoretical Intuition}\label{sec:Theoretical_Intuition}

As discussed in Section~\ref{sec:intro}, Artificial Intelligence is transforming nearly every aspect of modern life, reshaping industries, accelerating creativity, and redefining how humans interact with technology. At the core of these advancements lie generative models that are fundamentally dependent on the data used to train them. The quality, diversity, and originality of this data determine the richness and expressiveness of the model's output. However, with the rapid proliferation of AI-generated content, the composition of the internet is changing. Consequently, the training datasets that are scraped from it have begun to change. Increasingly, portions of these datasets contain synthetic data produced by earlier generations of generative models. When future models are trained on this self-generated, synthetic content, the diversity of the training corpus diminishes. Over time, the model repeatedly reinforces its own biases and stylistic patterns, rather than learning from genuinely novel data. This recursive process of training on self-produced data gradually erodes the information that was initially sparse and unique in the original dataset. As a result, subsequent generations of models begin to exhibit homogenized, repetitive, and less creative outputs. In essence, the model’s understanding of the data distribution becomes increasingly narrow, leading to a progressive loss of diversity and semantic integrity in its subsequent generations.

This intuition has been previously formalized in the literature by works such as  \cite{shumailov2024collapse} and \cite{dohmatob2024strongmodelcollapse}, which mathematically formalized how recursive self-training can cause informational degradation in generative systems. Building upon these foundations, we seek to visualize and quantify how similarity and diversity evolve within synthetically contaminated datasets.
\par
\definitionbox{Definition 3.1.1}{
\label{def:modelcollapse}
\textbf{Model collapse:} A degenerative process affecting successive generations of learned generative models, wherein the synthetic data produced by one generation contaminates the training corpus of subsequent generations, leading to a gradual degradation of diversity and semantic integrity in the model outputs.
}
This pollution affects the performance of the models in multiple ways.
\begin{itemize}
    \item \textbf{Increased Statistical Error}
    \subitem This phenomenon arises due to the finite sampling of data, which introduces discrepancies between the estimated feature distribution and the true underlying distribution. In other words, when only a limited subset of data is used to approximate the true data-generating process, small sampling errors can propagate and amplify across generations of models. 
    \subitem When synthetic data is introduced into these datasets, this discrepancy becomes more pronounced. Synthetic samples tend to \textit{oversample} the most likely feature values while \textit{undersampling} the less likely ones. Consequently, the estimated data distribution becomes increasingly concentrated around high-probability regions of the feature space. This shift amplifies the likelihood of already common patterns and suppresses rare or unique ones, resulting in more homogeneous and less diverse model outputs in subsequent generations. This is also referred to as \textit{"Forgetting the Tails of the Distribution"}\citep{shumailov2024collapse,dohmatob2024strongmodelcollapse}
    \item \textbf{Decreasing Generalization}
    \subitem As successive generations of large language and diffusion models continue to increase in scale, their sizes surpass those of previous generations by several orders of magnitude, incorporating billions of additional parameters at each stage, For Example, OpenAI models GPT-2\citep{gpt2} had about 1.5 billion data points in the largest model, it's successive generation, GPT-3\citep{gpt3} had about 175 billion parameters. Consequently, the risk of overfitting becomes more pronounced as the training data distributions grow increasingly homogeneous. In such cases, the effective information content of the training data may become equivalent to that of a much smaller yet more diverse dataset. When models of massive capacity are trained on such limited informational diversity, overfitting, which is the phenomenon characterized by a model’s inability to generalize effectively to novel or unseen data, is likely to occur.
    \item \textbf{Amplification of Biases}
    \subitem The open internet remains susceptible to misinformation, propaganda, and fabricated content. Consequently, the datasets scraped from online sources for training generative models are likely to contain inherent biases originating from such inaccuracies. Under typical circumstances, as factual information becomes more prevalent, the influence of falsified content tends to diminish over time. However, with the rapid proliferation of AI-generated material, models previously trained on biased data may inadvertently amplify these distortions. As such models generate and publish new content online, the proportion of synthetic data reflecting pre-existing biases can grow exponentially, further reinforcing and overrepresenting falsified information relative to unbiased content. This can be a part of \textit{Knowledge Collapse of AI models}\citep{Peterson_2025}
    
\end{itemize}
This study investigates the temporal evolution of diversity metrics in a dataset constructed from a fixed set of sources to construct a specific timeline for Model Collapse.

\section{Methodology}\label{sec:Data}
The methodology employed in this study follows a multi-stage approach designed to capture the temporal evolution of diversity metrics used to train language models. It consists of the following key stages:
\begin{enumerate}
    \item \textbf{Data Preprocessing:} Appropriate cleaning and filtering are applied to the raw data to remove inconsistencies and irrelevant entries. The processed data are then converted into structured formats that can be efficiently utilized in subsequent analytical steps.

    \item \textbf{Embedding:} Since computers cannot inherently comprehend the flow or contextual meaning of natural language, Natural Language Processing (NLP) models are employed to convert textual data from text or pages into vector embeddings. These embeddings encapsulate not only the semantic content but also the tone and writing style of the original text.

    \item \textbf{Compute a numerical metric for Diversity:} Use the vector embeddings computed in the previous state to compute a similarity or diversity metric between two elements within the dataset and the total average diversity in a database.

    \item \textbf{Visualization:} The resulting trends and relationships are visualized through a series of plots that illustrate the evolution of diversity metrics across years.
    \end{enumerate}
\subsection{Data Pre-processing}

The raw data extracted from the source database cannot be directly utilized for linguistic or textual analysis, as it typically contains inconsistencies in format, language, and content structure. Before any modeling stage, it is essential to ensure that the dataset is both linguistically and contextually homogeneous to prevent the introduction of external variance unrelated to the phenomenon, such as different writing styles for various languages, which can cause variability or diversity. Another example is how the tone of writing changes based on the audience. For instance, the tone of writing for a newspaper would be different from that of writing for a social media post, which again would cause variability that can mask the evolving recursive similarity. The effects and the need to filter data for linguistics and NLP are discussed in greater detail in \citep{NLPDataPreprocessing}

To avoid this, the preprocessing pipeline first standardizes the textual data to a uniform language domain. Multilingual or mixed-language entries are detected and either \textit{translated} to a common language or \textit{excluded} from the corpus to maintain consistency. This step minimizes linguistic noise that could otherwise mask or distort the patterns of similarity emerging from recursive training dynamics.

Next, the dataset is filtered based on source-type similarity. Data originating from distinct platforms or content categories (e.g., news articles, social media posts, and academic papers) exhibit inherently different stylistic and structural properties. 

If such heterogeneity is not controlled for, it may introduce what can be termed as \textit{irreducible diversity}—a baseline level of variation independent of the recursive generative contamination under study. Controlling for this factor ensures that any observed reduction in diversity over time is attributable to recursive model training rather than cross-domain variability.

These operations convert the cleaned corpus into a structured and machine-readable form suitable for subsequent embedding generation and diversity metric computation.

\subsection{Vector Embeddings}
The embeddings were generated using Transformer-based architectures. These models were chosen due to their documented ability \citep{Tranfomers.Ability, Transformers.Ability.Survey} to capture contextual relationships and semantic meaning within sentences. 

Each text input was passed through this Transformer model to obtain a high-dimensional vector representation. The output embeddings capture not only syntactic information but also semantic relationships, tone, and style of writing, enabling downstream models to perform context-aware tasks.
$$ x_{i,j,k}=f(\dataset[i])$$ where $f(.): D\to \reals^m$ is the transformer based encoder and $D$ is the set of all possible strings. 

\subsubsection{Transformers}
Transformers, introduced in \citep{vaswani2023attentionneed}, have fundamentally transformed the field of Artificial Intelligence through their significant contributions to Natural Language Processing (NLP) and the development of large language models. At the core of this architecture lies the self-attention mechanism, which enables the model to examine every word or token within an input sequence and determine the relative importance of each in relation to others. A Transformer is composed of multiple stacked layers of these attention blocks, where each layer progressively refines the model’s understanding of the input. Lower layers primarily capture syntactic structures such as grammar, whereas higher layers abstract deeper semantic representations, including meaning and intent. The foundational structure of a Transformer consists of two main components: an encoder and a decoder. However, in certain applications—such as text embedding and representation learning, only the encoder is employed to convert text into dense vector representations. In contrast, generative language models typically utilize both the encoder and decoder to produce coherent and contextually relevant sequences of words.

In this study, the encoder component of the Transformer model was employed to generate vector embeddings from textual data. By leveraging multi-head self-attention, the encoder effectively captured semantic and syntactic dependencies across sentences, producing context-rich and pattern-based representations suitable for downstream analysis.
\begin{figure}[H]
    \centering
    \includegraphics[width=0.3\linewidth]{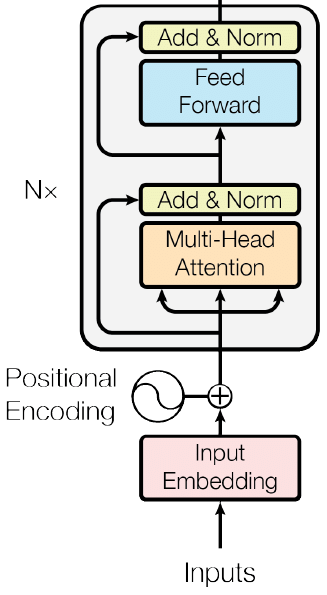}
    \caption{Schematic Representation of a Transformer Encoder "BERT", A multi-layer attention-based architecture, whose processing can be summarized as a sequence of Multi-Head Attention, Residual Addition and Normalization, and Feed-Forward layers repeated N times. This is also the transformer used for this work. Image sourced from\citep{castellucci2019multilingualintentdetectionslot} }
    \label{fig:transformer_encoder}
\end{figure}

\subsection{Similarity Analysis}
In this study, cosine distance is employed as the primary metric for quantifying semantic similarity between textual embeddings. The Cosine Distance uses the vector embeddings of the textual information generated by the transformer-based encoder. The Cosine Distance will be calculated as $$q_{i,j,k}=\frac{x_{i,j}^Tx_{i,k}}{||x_{i,j}||||x_{i,k}||}$$ where,\\ $||.||$ represents the 2-norm of any vector \\ Notation defined in Section \ref{sec:Problem_Statement}.

In this notation, $q_{i,j,k} \in [0,1]$, where
\begin{itemize}
    \item $q_{i,j,k}=0$ represents "zero" relevance between vectors $x_{i,j}$ and $x_{i,k}$ 
    \item $q_{i,j,k}=1$ represents "same contextual meaning" between vectors $x_{i,j}$ and $x_{i,k}$.
    \item $0<q_{i,j,k}<1$ represents partial relevance between the vectors $x_{i,j},x_{i,j}$, the greater this quantity, the greater the similarity between both these vectors.
\end{itemize}

Cosine distance has long been recognized as an effective measure for quantifying the relevance between two vector representations. It has been extensively applied in various domains, including information retrieval, search engine ranking, and query matching operations, such as those employed by Google Search to retrieve semantically relevant results. This widespread adoption underscores the suitability of cosine distance as a metric for assessing similarity in textual data. Furthermore, the effectiveness of cosine-based similarity has been the subject of extensive analysis and validation \citep{Cosine.Theory_1,Cosine.Theory_2,Cosine.Theory_3,Cosine.Theory_4,Cosine.Theory_5} within the academic community, establishing it as a reliable and interpretable measure for evaluating semantic relationships between embeddings.

The theoretical foundation for this choice lies in the attention mechanism that drives Transformer architectures. Both self-attention and cross-attention operate by comparing vector representations, referred to as queries and keys, to determine the relevance between tokens. This process is formalized as the scaled dot-product attention, as described in \cite{vaswani2023attentionneed}.

$$
Attention=softmax\left(\frac{\mathbf{QK^T}}{\sqrt{d_k}}\right)\mathbf{V}
$$
where,\\ $\mathbf{Q}$ is the query vector,\\ $\mathbf{K}$ is the key vector, \\ $\mathbf{V}$ is the value vector, \\$d_k$ is the dimension of $\mathbf{K}$

\subsubsection{Intuition}
The vector generating encoder models are trained within an unsupervised encoder–decoder configuration, with the objective of minimizing information loss between the input sequence and its reconstructed output. During training, the Transformer iteratively adjusts the parameters of its self-attention and cross-attention layers—namely, the query, key, and value matrices—so that semantically or contextually related tokens consistently produce higher attention scores. Over successive iterations, this optimization encourages words and phrases that frequently co-occur or share similar semantic roles to occupy proximal regions in the embedding space. Consequently, the decoder’s cross-attention modules, which also rely on cosine-based similarity, learn to recognize and attend to related information encoded in nearby vector directions, thereby reducing the difference between the input and the reconstructed output—even when expressed through different linguistic forms. Conversely, unrelated terms diverge spatially, reinforcing a geometric organization of linguistic relevance within the learned representation space.

\section{Experiment}\label{sec:experiment}
This section presents the practical implementation of the proposed similarity analysis framework. Textual data were encoded using a Transformer-based model to generate high-dimensional embeddings that capture both contextual and semantic information. Cosine distance was then applied to these embeddings to quantify the degree of similarity between text pairs. The analysis aims to validate whether the similarity within the data increases as a result of growing pollution in datasets caused by the presence of AI-generated content. The code to reproduce this is publicly available on GitHub, referenced. The code was executed on a machine equipped with an Intel Ultra 9 155H Processor, aided by a CUDA-enabled Nvidia RTX 4070 graphics processor with 32 GB of memory.
\subsection{Dataset and Preprocessing}
For this study, the Common Crawl database\citep{cc:2010} was selected, because it provides year-wise corpora, allowing us to track the evolution of similarity metrics over time and analyze the potential timeline of AI model collapse. The Common Crawl dataset is a large-scale web crawl of publicly available websites, hosted on AWS, and distributed as wget and index files that can be easily processed using various Python libraries.

During preprocessing, it is essential to acknowledge that the Common Crawl contains content in multiple languages and from a diverse set of websites. As discussed in the previous section, such diversity can obscure the emerging patterns of similarity that result from the increasing presence of AI-generated content. To mitigate this masking effect, we filter the dataset to include only English-language websites and further isolate articles from Wikipedia \citep{wikipedia2025}. This restriction ensures that the data primarily originates from a consistent group of human authors, thereby reducing but not eliminating \textit{irreducible diversity}, helping prevent it from overshadowing the developing similarity trends.

In this study, source-based filtering was performed by isolating crawl paths that contained \textit{“wikipedia.com”} as the domain. Language-based filtering was subsequently applied using the pycld2 library \citep{pycld2} to exclude all non-English data. Since domain-based filtering is computationally less intensive than language-based filtering, the algorithm first extracted articles originating from the Wikipedia domain and then verified their language. To further minimize computational overhead, the language detection process was limited to the first 2000 words of each webpage. If these initial words were identified as English, the entire document was assumed to be written in English.

\subsection{Vector Embeddings and Cosine Similarity}
The Vector embeddings were created by a pretrained Transformer model imported through sentence-transformer from HuggingFace \citep{wolf2020huggingfacestransformersstateoftheartnatural}. The pre-trained weights of \textit{"BAAI/bge-large-en-v1.5"} \citep{bge_embedding}, a transformer-based embedding model optimized for \textit{semantic similarity} and retrieval tasks. The model projects each text input into a \textbf{1024-dimensional embedding space}, where contextual and semantic information is preserved geometrically, as shown with transformer-based models in Section \ref{sec:Data}. It is trained using contrastive learning, aligning semantically similar sentence pairs while pushing dissimilar pairs apart. The resulting embeddings exhibit strong clustering behavior for texts with shared meaning, tone, or origin. This property is particularly suitable for identifying semantic redundancy and potential AI-generated content through similarity. The use of \textit{bge-large-en-v1.5} ensures sensitivity to nuanced contextual relationships while maintaining robustness against surface-level lexical variations, thereby enabling the reliable detection of similarity patterns indicative of AI pollution in text datasets. The model and encoding process were run on a GPU to take advantage of parallel computing through CUDA.

Each dataset entry is transformed into a fixed-length vector representation using the encoder, after which cosine similarity ($q_{i,j,k}$) is computed pairwise across embeddings to quantify semantic overlap using NumPy\citep{harris2020array}. Then the average similarity is calculated for generalizing the whole database ($q_{i}$).

\subsection{Results and Discussion}\label{sec:Discussion}
\begin{figure}
    \centering
    \pgfplotstableread[col sep=comma]{cosine_similarity.csv}\similaritydata
    \begin{tikzpicture}
        \begin{axis}[
            width=1*\textwidth,
            height=8cm,
            xlabel={Year},
            ylabel={Average Cosine Similarity},
            xmin=2013, xmax=2025,
            grid=major,
            grid style={dashed,gray!30},
            tick label style={font=\small},
            label style={font=\small},
            legend style={font=\small,at={(0.02,0.98)},anchor= north west},
            xticklabel style={/pgf/number format/.cd, set thousands separator={}, fixed, precision=0},
        ]
            \addplot[
                color=blue,
                thick,
                mark=*,
                smooth,
                color = black
            ] table[
                x=year,
                y=cosine similarity
            ] {\similaritydata};
            \legend{\small{Average Similarity}}

            \addplot[
            domain=2013:2025,
            samples=200,
            dashed,
            color=blue,
        ]
        {0.35 + 0.0935*(1 - exp(-0.1029*(x - 2013)))};
        \addlegendentry{\small{Fit Model $0.35 + 0.0935(1 - e^{-0.1029(x - 2013)})$}}
        \end{axis}
    \end{tikzpicture}
    \caption{Average cosine similarity from 2013 to 2025 showing an increasing trend in homogeneity driven by Synthetic AI generated textual data, that has significantly increased after the 2017 breakthrough in Natural Language processing using transformers and then the public adoption of Transformer based LLMs like ChatGPT API \citep{gpt3,gpt2} in late 2022.}
    \label{fig:similarity}
\end{figure}
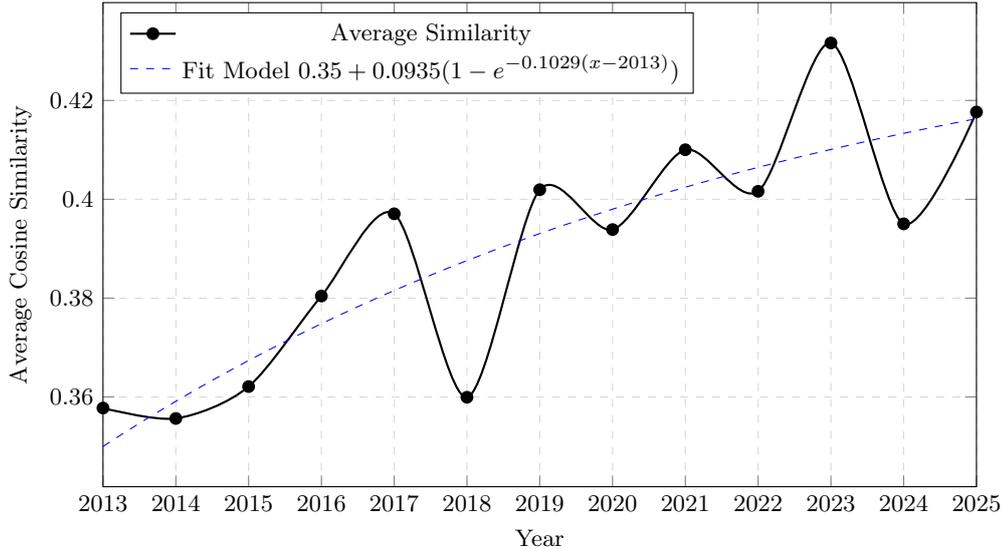

After computing the average similarity values, the results were visualized to better illustrate the observed trends. Figure~\ref{fig:similarity} presents the variation of average similarity across different time periods, highlighting the gradual increase in textual homogeneity within the datasets analyzed.

Figure~\ref{fig:similarity} presents a clear upward trend in the similarity measurements, indicating a gradual yet consistent increase in the homogeneity of textual data over time. This trend suggests that the linguistic and semantic diversity within large-scale text corpora is diminishing. Such a pattern is strongly correlated with the rapid proliferation of AI-generated content in the public web domain. Notably, a significant increase can be observed around the year 2019-2021, the period marking the introduction of transformer architecture, development of transformer-based language models, and public release and widespread adoption of Large Language Models (LLMs) with GPT-2 and GPT-3 released during 2019 and 2021 \citep{gpt2,gpt3}. 

A noticeable increase in similarity can also be observed even before the widespread public adoption of Large Language Models (LLMs). This early rise can be attributed to the use of earlier neural architectures such as Long Short-Term Memory (LSTM) networks, Recurrent Neural Networks (RNNs), and other language models that were deployed by various websites for tasks like translation, summarization, and text normalization. However, since these technologies were not adopted on a large scale, their overall impact on global linguistic homogeneity remained relatively limited.

The acceleration of homogeneity beyond this point implies that the increasing prevalence of AI-generated text is influencing the natural variability of online content. This phenomenon aligns closely with the \textit{Forgetting the Tails} effect described in prior theoretical work~\citep{dohmatob2024strongmodelcollapse, shumailov2024collapse}, wherein iterative training on AI-influenced data leads to a collapse of diversity and the gradual erosion of rare or unique features within the data distribution. This also leads to a reduction in the ratio of parameters to that of the equivalent number of data points if all data points were diverse, leading to an overfit model that might fail if given unseen and novel problems and data. 

For the estimation of the projected timeline, we fit the obtained empirical data to an exponential growth-based function, denoted as \( y \mapsto h(y): \mathbb{R} \to \mathbb{R} \), where \( y \in \mathbb{R} \) represents the input year and \( h(y) \) corresponds to the estimated data diversity or similarity metric. The parametric form of this function is defined as

\begin{equation*}
    h(y) = h_0 + a\left(1 - e^{-b(y - y_0)}\right),
\end{equation*}

Where \( h_0 \) and \( y_0 \) represent the approximate baseline values for data diversity and the starting year, respectively. While these can also be treated as free parameters, they are fixed in this analysis to reduce model complexity, given the limited availability of historical data for robust generalization. The remaining parameters, \( a \) and \( b \), are determined empirically based on the dataset.

The parameters were estimated using gradient descent optimization\citep{DiscoveryLab.ShiIyengar2020} implemented via the \texttt{SciPy} library~\citep{Virtanen_2020}, by minimizing the Euclidean loss function:

\begin{equation*}
    \mathcal{L} = \sum_{D} \left(q_i - h(y_i)\right)^2,
\end{equation*}

where \( q_i \) represents the observed similarity and \( h(y_i) \) is the model prediction for the corresponding year \( y_i \).

Using the available data, the fitted function is obtained as:
\begin{equation*}
    h(y) = 0.35 + 0.0935\left(1 - e^{-0.1029(y - 2013)}\right).
\end{equation*}

This function models the gradual increase in average similarity over time, capturing the observable rise in homogeneity across datasets. Based on this fitted model, the projected years corresponding to 90\%, 95\%, and 99\% saturation levels are summarized in Table~\ref{tab:saturation_years}.

\textit{Remark:} For x\% saturation, we calculate the possible y for which $1 - e^{-0.1029(y - 2013)}=\frac{x}{100}$ \\
\begin{table}[ht]
\centering
\label{tab:saturation_years}
\begin{tabular}{|l|l|}
\hline
\textbf{Saturation Level} & \textbf{Year} \\
\hline
90\% & 2035 \\
95\% & 2042 \\
99\% & 2057 \\
\hline
\end{tabular}
\caption{Estimated years at which the model reaches different saturation levels of similarity.}
\end{table}
Therefore, at this level and pattern of generation, AI model collapse is predicted to occur after 2035.
\subsection{Limitations}
This estimation should be interpreted as a preliminary projection based on currently available data. It does not account for the potential acceleration in data generation resulting from the development and widespread adoption of more powerful future AI systems. The formulation of increasingly capable foundation models, multimodal architectures, or other disruptive technologies could dramatically alter the rate at which synthetic data is produced (Similar the captured rise in rate of generation seen after the introduction of transformers by \cite{vaswani2023attentionneed} in 2017-2021), thereby shifting the projected timeline for model saturation or collapse. Furthermore, given that the large-scale public deployment of generative AI models began only in recent years, the amount of empirically verifiable data remains limited. Consequently, this analysis should be viewed as an initial approximation that captures the present observable trends rather than a definitive prediction. As the field evolves and richer datasets become available, subsequent research will be necessary to refine these estimates and assess the long-term trajectory of linguistic and semantic homogeneity in training corpora.

It is also important to note that this early estimate is inherently noisy due to the presence of irreducible diversity in natural language data. Variations in the number of samples within each year’s corpus can influence the degree to which this diversity affects the computed similarity. As a result, minor fluctuations in the similarity trend—visible in the figure—may be attributed to these random sampling effects rather than underlying semantic convergence.

The empirical evidence demonstrated in the figure thus substantiates these theoretical predictions, showing that the data currently being used for training contemporary models already exhibit measurable signs of this collapse. In essence, the observed rise in similarity metrics provides a tangible reflection of how the feedback loop between AI-generated content and training datasets may be driving the ecosystem toward increasing uniformity and reduced informational richness.
\section{Conclusion}\label{sec:Conclusion}
This study presents an empirical framework for analyzing the evolution of textual similarity in large-scale web corpora, using the filtered Common Crawl dataset, as shown in Section \ref{sec:experiment}, as a proxy for any database. By encoding text through Transformer-based embeddings and applying cosine similarity as a measure of homogeneity, the analysis revealed a consistent upward trend in linguistic and semantic uniformity over the years. This finding aligns with recent theoretical predictions \citep{dohmatob2024strongmodelcollapse, shumailov2024collapse} of \emph{Model Collapse}, wherein successive generations of AI models trained on synthetic data exhibit reduced diversity and information richness.

The data-based exponential model indicates that average similarity has been steadily increasing since 2013, with a significant acceleration from 2018 to 2022, coinciding with the public release of large language models by OpenAI, the GPT-2\citep{gpt2} and GPT-3\citep{gpt3} models. Extrapolating from the approximation suggests that the ecosystem may reach 90\%, 95\%, and 99\% saturation around 2035, 2042, and 2058, respectively. These milestones represent critical thresholds beyond which the dominance of AI-generated content could begin to self-reinforce, diminishing the diversity of future training datasets, and estimating the phenomenon of model collapse to occur at a 90\% or greater similarity rate, which means the approximate time of collapse of AI models will be \emph{2035}

The study also emphasizes that these projections are preliminary. The estimates do not yet account for the rapidly evolving landscape of generative technologies or the emergence of new modalities, which may either exacerbate or mitigate data homogenization. Furthermore, the limited temporal depth of currently available data constrains the precision of long-term predictions. Future research should extend this work by integrating multi-modal datasets, analyzing the effects of current techniques to mitigate this phenomenon, such as adaptive feedback between human and synthetic content generation, and developing mechanisms to preserve data diversity in large-scale AI ecosystems.

In summary, this study provides an early quantitative signal supporting theoretical concerns about recursive data pollution and the long-term sustainability of generative AI. Continuous monitoring of data diversity metrics, coupled with responsible dataset curation, will be essential to ensure that AI development remains grounded in the creative and heterogeneous essence of human knowledge.

\section*{Acknowledgment}
The authors thank the United States Army and, National Science Foundation for their support in these preliminary investigations resulting in the analysis of this study. The authors also thank their parents, colleagues, and various other students and faculty for their suggestions.

\small{
\bibliography{references,discoverylab_master,ccref}
}
\end{document}